# BIOMETRIC AUTHORIZATION SYSTEM USING GAIT BIOMETRY

L.R Sudha[1], Dr. R. Bhavani[2]

Department of CSE, Annamalai University, Chidambaram, TamilNadu, India

[1]sudhaselvin@ymail.com, [2]shahana_1992@yahoo.co.in

## ABSTRACT

*Human gait, which is a new biometric aimed to recognize individuals by the way they walk have come to play an increasingly important role in visual surveillance applications. In this paper a novel hybrid holistic approach is proposed to show how behavioural walking characteristics can be used to recognize unauthorized and suspicious persons when they enter a surveillance area. Initially background is modelled from the input video captured from cameras deployed for security and the foreground moving object in the individual frames are segmented using the background subtraction algorithm. Then gait representing spatial, temporal and wavelet components are extracted and fused for training and testing multi class support vector machine models (SVM). The proposed system is evaluated using side view videos of NLPR database. The experimental results demonstrate that the proposed system achieves a pleasing recognition rate and also the results indicate that the classification ability of SVM with Radial Basis Function (RBF) is better than with other kernel functions.*

## 1. INTRODUCTION

A wide variety of systems requires reliable personal recognition schemes to either confirm or determine the identity of an individual requesting their services. The purpose of such schemes is to ensure that the rendered services are accessed only by a legitimate user and no one else. Some examples of such applications include secure access to buildings, computer systems, and ATMs. In the absence of robust personal recognition schemes, these systems are vulnerable to the wiles of an impostor. Biometric recognition or, simply, biometrics refers to the automatic recognition of individuals based on their physiological and/or behavioural characteristics. By using biometrics, it is possible to confirm or establish an individual's identity based on "who she is" rather than by "what she possesses" (e.g., an ID card) or "what she remembers" (e.g., a password).

The interest in gait as a biometric is strongly motivated by the need for an automated recognition system for visual surveillance and monitoring applications. Recently the deployment of gait as a biometric for people identification in surveillance applications has attracted researches from computer vision. The development in this research area is being propelled by the increased availability of inexpensive computing power and image sensors. The suitability of gait recognition for surveillance systems emerge from the fact that gait can be perceived from a distance as well as its non-invasive nature. Although gait recognition is still a new biometric, it overcomes most of the limitation that other biometrics suffer from such as face, fingerprints and iris recognition which can be obscured in most situations where serious crimes are involved. As stated above, gait has many advantages, especially unobtrusive identification at a distance, so that unauthorized and suspicious persons can be recognized when they enter a surveillance area, and night vision capability which is an important component in surveillance. This makes gait a very attractive biometric for real time monitoring as well as for access control at sites of high risk.

Current gait recognition methods may be mainly classified into two major categories, namely model-based and motion-based methods. Usually, in the model-based methods, the human body structure or motion is modelled first, and then the image features are extracted by the measure of structural components of models or by the motion trajectories of body parts [1, 2]. Most existing motion-based approaches can be further divided into two main classes, state-space methods and spatiotemporal methods [3, 4]. The state-space methods consider gait motion to be composed of a sequence of static body poses, and recognize it by considering temporal variations observations with respect to those static pose. The spatiotemporal method characterizes the spatiotemporal distribution by collapsing the entire 3D spatiotemporal (XYT) data over an entire sequence into a terse 1D or 2D signal(s).

In [5] a baseline algorithm using raw silhouette for human identification was proposed. In [6], the authors used outer contour and unwrapped it into a distance signal to recognize a person. Yang et al. [7] obtained the dynamic region in gait energy image (GEI) which reflects the walking manner of an individual, called enhanced GEI. In [8], Hong et al. extracted a gait feature called mass vector from the number of pixels with nonzero value in a given row of the binary silhouette. They used dynamic time-warping approach to match gait templates formed by the mass vectors. Xu et al. applied Marginal Fisher Analysis (MFA) for dimensionality reduction to achieve a compact gait representation in [9]. They also modified the MFA to matrix based version to process 2-D input directly in the form of gray-level averaged images. In [10] an angular transform was introduced and was shown to achieve very encouraging results. In [11] a new system based on radon transform and LDA was proposed. In [12], the authors used a vector data scanned in horizontal, vertical and diagonal direction to represent the gait. Guo and Nixon [13] presented an efficient solution based on mutual information (MI) to select important features for gait recognition. These early results further confirm that gait has a rich potential for human identification. Based on these considerations a new gait representation scheme is proposed and is described in the next section.

## 2. PROPOSED APPROACH

There are many properties of gait that might serve as recognition features. Early medical studies suggest that there are 24 different components to human gait and if all movements are considered, gait is unique [14]. However from a computational perspective, it is quite difficult to accurately extract some of the components such as angular displacements of thigh, leg and foot using current computer vision system, and some others are not consistent over time for the same person. So precise extraction of body parts and joint angles in real visual imagery is a very cumbersome task as non-rigid human motion encompasses a wide range of possible motion transformations due to the highly flexible structure of the human body and to self occlusion. Furthermore clothing type, segmentation errors and different viewpoints post a substantial challenge for accurate joint localization. Hence, the problem of representing and recognizing gait turns out to be a challenging one. In this paper we developed a hybrid holistic approach which is computationally affordable for real-time applications.

In this paper we are concerned with only side view videos and normal gait. This is because gait of a person is easily recognizable when extracted from side view of the person and majority of the people have normal walk. A careful analysis of gait reveals that it has two important components, a structural component which captures the physical build of a person and a dynamic component which captures the transitions that the body undergoes, during a walk cycle. We categorize them as spatial components, temporal components and wavelet features.

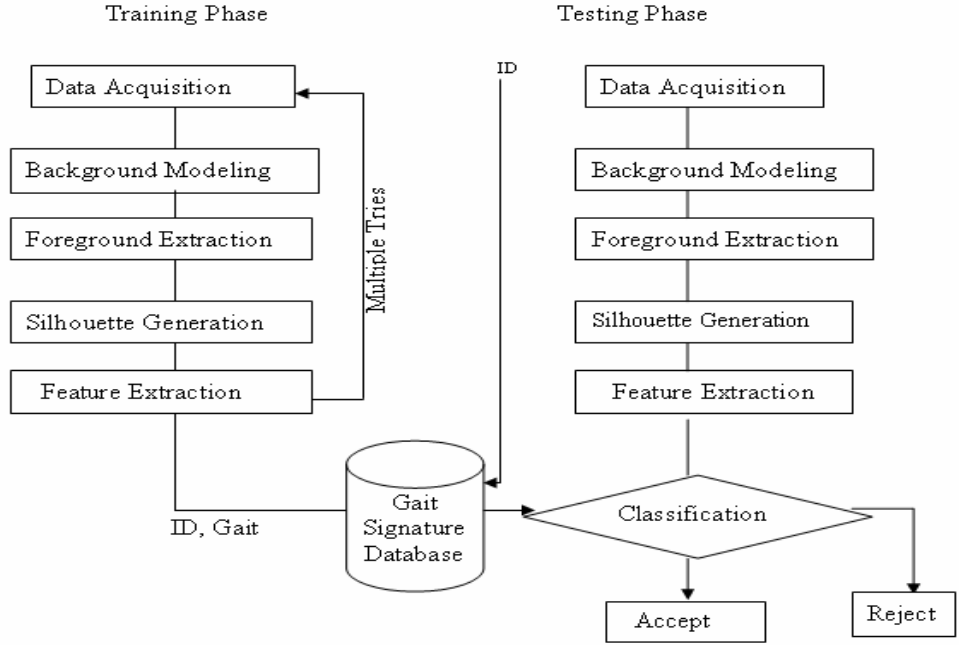

Figure 1. Human Gait Recognition System

Our Biometric Authorization system is schematically shown in Figure 1. The input to our system is a gait video sequence captured by a static camera. Once the video is captured, binary silhouettes of the walker are generated by using a background subtraction process which includes two important steps background modelling and foreground extraction. Then the features are extracted as numerical information. All the above steps are repeated for multiple video sequences and mean of feature values are stored in the database along with the identity of the walker. Then pattern recognition classifiers are trained with the created database in training phase. During testing phase, binary silhouette of the test video is generated as in training phase and the individual is recognized by comparing the obtained features with the ones previously stored in the database by the classifier.

The remainder of this paper is organized as follows. Section 3 describes the modules of our system, section 4 provides experimental results and section 5 contains conclusion.

## 3. MODULES DESCRIPTION

Our algorithm consists of five basic modules Background Modelling, Foreground Extraction, Silhouette Generation, Feature extraction and Recognition which are described in the subsections below.

### 3.1 Background Modelling

Extracting moving objects from a video sequence captured using a static camera is a fundamental and critical task in visual surveillance. A common approach to this critical task is to first perform background modelling to yield a reference model. By using the sequence of frames obtained from video, background can be modelled which can be later used to visualise moving objects in the video. Much research has been devoted to develop a background model that is robust and sensitive enough to identify moving object of interest.

In our implementation, for background modelling we make the fundamental assumption that the background will remain stationary. This necessitates that the camera be fixed and that lighting does not change suddenly. It is also possible to achieve accurate segmentation without this assumption, but such generality would require more computationally expensive algorithms.

We have compared three different background modelling techniques which are widely used in many works, namely Change Detection Mask (CDM), Using median value, and Histogram based in order to choose simple but efficient technique. These three techniques are discussed briefly below and the experimental results are tabulated.

### 3.1.1 Change Detection Mask technique (CDM)

A basic change detection algorithm takes image sequence as input and generates a new image called a *change mask* that identifies changed regions [15]. Let $I_i(x,y)$ represent a sequence including N images, i represents the frame index ranging from 1~N. The value of CDM can be computed by

$$CDM_i(x, y) = \begin{cases} d, & if\ d >= T \\ 0, & else \end{cases}. \quad (1)$$

$$d = |I_{i+1}(x, y) - I_i(x, y)|. \quad (2)$$

Where T is the threshold value determined through adaptive thresholding. $CDM_i(x,y)$ represents the value of the brightness change in pixel position along the time axis. The value in pixel point (x,y) is estimated by the value of median frame of the longest labelled section in the same pixel point (x,y).

### 3.1.2. Using Median value

An estimate of the background image can be obtained by computing the median value for each pixel in the whole sequence [16]. Let B(x, y) is the background value for a pixel location (x, y), med represents the median value, [I(x, y, t),....I(x, y, t+n)] represents a sequence of frames.

$$B(x, y) = med[I(x, y, t),...I(x, y, t + n)]. \quad (3)$$

Use of the median relies on an assumption that the background at every pixel will be visible more than fifty percent of the time during the training sequence. The output of this typically will contain large errors when this assumption is false. The advantage of using median is that it avoids blending pixel values.

### 3.1.3. Histogram Based Background Modeling

In this type of modeling, background value for each pixel position is based on the dominant bins intensity value based on histogram analysis. Illumination changes and camera jitter will greatly affect the performance of the algorithm.

Figure.2 and Table I shows the generated background models and the modelling time by the above mentioned three techniques. We found that though histogram based computation is comparatively faster than others, it does not model the background clearly for the second and third input, Gait video2 and Gait video3. As median value calculation is less complex and takes very less time when compared with CDM, in our biometric system we used median value to model the background.

Table 1  Processing Time For Background Modelling Techniques In Seconds

| Background Model | Gait Video 1 | Gait Video 2 | Gait Video 3 |
|---|---|---|---|
| Change Detection Method | 9.1878 | 177.5332 | 35.8499 |
| Median value | 1.0361 | 19.0566 | 6.0849 |
| Histogram based | 0.4894 | 4.7969 | 1.1254 |

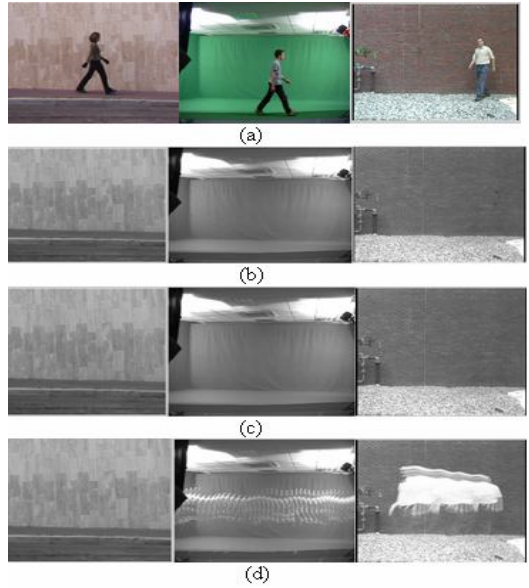

Figure 2. (a) Input Videos (b), (c) and (d) background models generated through CDM, Median and Histogram based techniques respectively.

### 3.2. Foreground Extraction

After background modelling, identifying moving objects from a video sequence is a fundamental and critical task in many computer–vision applications. In this paper performance of popular foreground extraction technique namely Frame difference is tested with the background models generated by the three background modelling techniques discussed above.

Frame Difference Method is a very simple method with only one operation, a image subtraction. In this method a pixel is marked as foreground if

$$|I_i(x, y) - B(x, y)| > T \qquad (4)$$

where T is adaptive threshold. From the performance results shown in Table 2, it is evident that frame difference with median value is better than others.

Table 2.
Processing time for Frame Difference technique with three different background models in seconds

| Background Model | Gait Video1 | Gait Video 2 | Gait Video3 |
|---|---|---|---|
| Change Detection Method | 13.7881 | 233.3640 | 41.2510 |
| Median value | 5.1341 | 31.9462 | 10.7328 |
| Histogram based | 4.2081 | 17.1799 | 6.7801 |

### 3.3. Silhouette Generation

In order to be robust to changes of clothing and illumination it is reasonable to consider the binarized silhouette of the object as shown in Figure 3. This can be obtained by thresholding the foreground image with a suitable threshold value T.

$$FG_k(x, y) = \begin{cases} 0 \; Background & if \; FG_k(x, y) > T \\ 1 \; Foreground & else \end{cases}. \quad (5)$$

The primary assumption made here is that the camera is static, and the only moving object in video sequences is the walker.

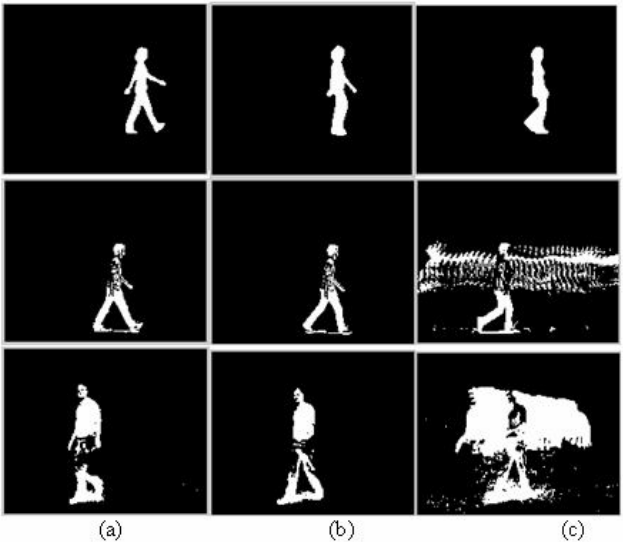

Figure 3. Silhouette Image in a) CDM b) Median c) Histogram based.

### 3.4. Feature Extraction

Before the process of feature extraction, we first place a bounding box around the binary silhouette to estimate the gait cycle and thereby to estimate the number of frames in a gait cycle. Knowing the number of frames in a gait cycle, the sequence of binary silhouettes is divided into sub sequences of gait cycle length. We then consider frames in two gait cycles in order to extract the characteristic feature vector. This reduces the processing time considerably. The process of gait cycle estimation is briefly described below.

### 3.4.1. Gait Cycle Estimation

Human gait is treated as a periodic activity within each gait cycle. A single gait cycle can be regarded as the time between two identical events during the human walking and usually measured from heel strike to heel strike of one leg. The computation of gait period and cycle partitioning are crucially important steps for gait recognition algorithms. Several methods have been proposed to estimate gait periodicity as it provides essential information for the extraction of gait features. To estimate the gait cycle the aspect ratio of the silhouettes bounding box is used in [17]. In our method width of the bounding box is used as it is periodic and the bounding box will be larger and shorter when the legs are farthest apart and thinner and longer when the legs are together.

### 3.4.2 Spatial Component Computation

Bounding rectangle's mean height, mean width, mean angle and mean aspect ratio are considered as spatial components. Bounding rectangle's mean height H is the representative height value for a person [18]. It is obtained by averaging the difference between upper and bottom points on y axis at each time instant t given by

$$H = \frac{\sum_{i=1}^{N}[Y_{up}(t) - Y_{bp}(t)]_i}{N} \quad (6)$$

where $Y_{up}$ is the upper point on y axis, $Y_{bp}$ is the bottom point on y axis and N is the number of frames in the gait cycle. Similarly Bounding rectangle's mean width W is the representative width value of a person. It is obtained by averaging the difference between right and left points on axis x at each time instant t.

Bounding rectangle's mean angle A is the representative diagonal value of a person. It is obtained by averaging the difference in degrees between axis x and the diagonal at each time instant t and aspect ratio AR can be obtained by dividing Height H by Width W.

### 3.4.3. Temporal Component Computation

Stride length, Step length, Cadence and Velocity are considered as temporal components. Stride length is the distance travelled by a person during one stride(or cycle) and can be measured as the length between the heels from one heel strike to the next heel strike on the same side. Two step lengths (left plus right) make one stride length. Step length and stride lengths are computed by finding the number of frames in a step and stride. Cadence is number of steps /minute and velocity is calculated by the equation given below.

$$Velocity = strideleng\,th \times 0.5 cadence \quad (7)$$

### 3.4.4. Wavelet Energy Component Computation

Studies have shown that human gait is quasi periodic and there are slight changes in the fundamental frequency and amplitude over time. Discrete Wavelet Transform (DWT) is used to perform such analysis due to its robustness against rotation, translation and scaling. Because of its great time and frequency localization ability, DWT can reveal the local characteristics of the input patterns, enabling good representation of the local features of the patterns. From this point of view, the wavelet descriptors are better than Fourier, since they are able to catch small differences between patterns.

Among the various wavelet bases, the Haar wavelet is the shortest and simplest and it provides satisfactory localization of signal characteristics in time domain. Therefore, Haar wavelet was chosen as the mother wavelet in this work. The silhouette area of each frame is decomposed by 2-D DWT using Haar wavelet kernel. From the low frequency subband we got one coefficient and from the detailed frequency subband we got two coefficients. Then mean and standard deviation is calculated as in eq. 8 and eq. 9 for all energy coefficients of each subband. Therefore the dimension of wavelet feature is six.

$$\mu = \frac{1}{N} \sum_{i=1}^{N} a_i. \quad (8)$$

$$\sigma = \sqrt{\frac{1}{N-1} \sum_{i=1}^{N} (a_i - \mu)^2}. \quad (9)$$

where $\mu$ is mean, $\sigma$ is standard deviation, N is number of frames in the gait cycle, and $a_i$ represents the different wavelet coefficients.

### 3.5. Human Recognition

Gait Recognition is a traditional pattern classification problem which can be solved by calculating the similarities between instances in the training database and test database. Let the spatial feature vector be S, the temporal feature vector be T and the wavelet feature vector be W. By fusing all these features at feature level, the feature vector H is represented as in eq. 10.

$$H = [S, T, W]. \quad (10)$$

This H is used to train SVM models to classify human gait.

Recently, Support Vector Machines (SVM), proposed by Cortes and Vapnik in 1995 has emerged as a powerful supervised learning tool for general purpose pattern recognition. It has been applied to classification and regression problems with exceptionally good performance on a range of binary classification tasks [19]. The primary advantage of SVM is its ability to minimize both structural and empirical risk leading to better generalization for new data classification even when the dimension of input data is high with limited training dataset. SVM can be trained both as a binary classifier and multi-class classifier. Number of kernels such as linear, polynomial and Gaussian radial basis function can also be used in SVM models. So far, no analytical or empirical study was conclusively established the superiority of one kernel over another.

## 4. EXPERIMENTAL RESULTS AND ANALYSIS

We have used strong computing software called Matlab to develop our work because Matlab provides image Acquisition and Image Processing Toolboxes which facilitate us in creating a good GUI and an excellent code.

### 4.1. Data Acquisition

The experimentation of the proposed gait recognition system is performed with images publicly available in the National Laboratory of Pattern Recognition gait database. It contains 240 sequences from 20 different subjects and four sequences per subjects in three different views. The properties of the images are: 24-bit full colour, capturing rate of 25 frames per second and the original resolution is 352 × 240. The length of each sequence varies with the time each person takes to traverse the field of view.

## 4.2. Results and analysis

For each side view videos of NLPR gait database, we first generate silhouette images using background subtraction algorithm and then spatial, temporal and wavelet features are extracted in the manner described in section 3. Then we trained SVM classifier by the feature vector H, and the gaits are classified by the trained models at last.

We first compare the performance rate (Accuracy) of spatial, temporal & wavelet features separately. Then the performance of different combinations of features types is compared. We see that in our experiments, the recognition rate is found to be increased when all the three feature types are fused together and it is shown in Table 3 and Figure 4. We also found that though temporal features gives poor performance while using separately, improves performance rate when fused with other features.

Table 3

Performance Comparison of Different Feature Types

| Feature Type | No. of Features | % of Performance |
| --- | --- | --- |
| Spatial Features (A) | 4 | 72.91 |
| Temporal Features (B) | 4 | 31.25 |
| Wavelet Features (C) | 6 | 91.66 |
| Spatial & Temporal Features (D) | 8 | 79.17 |
| Spatial & Wavelet Features (E) | 10 | 85.43 |
| Spatial, Temporal & Wavelet Features (F) | 14 | 97.91 |

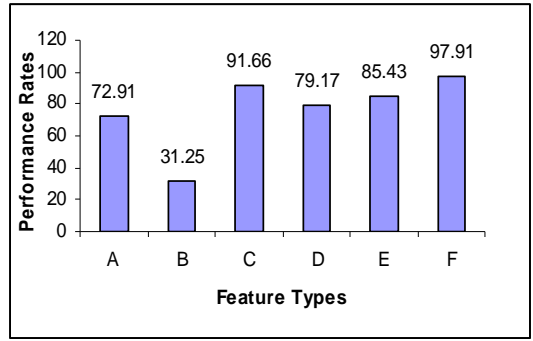

Figure 4. Performance rate of all feature types

In SVM, experiments were conducted with three kernel types for various values of regularization parameter (c) and other parameters such as degree of polynomial (d) in Polynomial Kernel and width of RBF function (σ). We found that classification performance of SVM depends on the selection of regularization parameter because it is the penalty parameter for misclassification. So it has to be carefully selected to achieve maximum classification accuracy. When compared across different kernels with various parameter values, Radial Basis Function (RBF) was found to perform the best in recognizing gait patterns. Best classification outcomes for different kernels are represented using accuracy rates and are tabulated in Table 4.

Table 4.

Performance rate for various SVM kernels

| SVM Kernel types | % of Performance |
|---|---|
| Linear | 77.08 |
| Polynomial | 77.08 |
| Radial Basis function | 97.91 |

We compared the proposed method with other well cited gait recognition approaches using NLPR gait database and is given in Table 5. From the table it is clearly seen that the proposed approach gain a better performance rate.

Table 5.
Comparison with State _of_ the Art Algorithms on NLPR database in the Canonical View

| Methods | Performance Rate (%) |
|---|---|
| Collins (2002) | 71.25 |
| BenAbedelkader (2002) | 82.50 |
| Phillips (2002) | 78.75 |
| Wang (2003) | 88.75 |
| Su (2006) | 89.31 |
| Lu (2006) | 82.50 |
| Geng (2007) | 90.00 |
| Bo Ye (2007) | 88.75 |
| Proposed Method | 97.91 |

In this work apart from Performance rate, other measures such as Precision, Recall, and F_measure which are more appropriate for comparison are also considered and the values are tabulated in Table 5. The formulas for the above performance measures are given below.

$$\Pr ecision = \frac{TP}{TP+FP} \tag{11}$$

$$\text{Re} call = \frac{TP}{TP+FN} \tag{12}$$

$$F\_Measure = \frac{2 \times \Pr ecision \times \text{Re} call}{\Pr ecision + \text{Re} call} \tag{13}$$

where TP is True Positive, FN is False Negative and FP is False Positive. These measures are calculated using confusion matrix of classification.

Table 6.
Comparison of Different Performance Measures

| % of Performance Measures | |
|---|---|
| Accuracy | 97.91% |
| Precision | 98% |
| Recall | 98% |
| F-Measure | 98% |

## 5. CONCLUSION

With mounting demands for visual surveillance systems, human identification at a distance has recently emerged as an area of significant interest. Gait is being considered as an impending behavioral feature and many allied studies have illustrated that it can be used as a valuable biometric feature for human recognition. The proposed method can effectively capture the gait characteristics for a side view. But in realistic surveillance scenarios, however it is unreasonable to assume that a person could always present a side view to the camera and hence the algorithm need to be extended to work in a situation where the person walks at an arbitrary angle to the camera.

**Authors**

L.R Sudha received her B.E degree in Computer Science and Engineering from Madras University, Chennai, India in 1991 and M.E degree in Computer Science and Engineering from Annamalai University, Chidambaram, India in 2007.

She is currently working as Assistant Professor and working towards the Ph.D degree. Her research work is focused on Video and Image Processing, Pattern Recognition, Computer Vision, Human Gait analysis and their applications in Biometrics. She has published 10 papers in International and National conference proceedings.

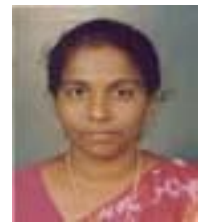

**Dr. R. Bhavani** received her B. E degree in Computer Science and Engineering in the year 1989 and the M.E degree in Computer Science and Engineering in the year 1992 from Regional Engineering College, Trichy. She received her Ph.D degree in Computer Science and Engineering from Annamalai University, Chidambaram, in the year 2007.

She worked in Mookambigai college of Engineering, Keeranur from 1990 to 1994, and she is now working as Associate Professor in Annamalai University, since 1994. She published 15 papers in international conferences and journals. Her research interest includes Image processing, Image Segmentation, Image Compression, Image Classification, Steganography, Pattern Classification, Medical Imaging, Content Based Image Retrieval and Software metrics.

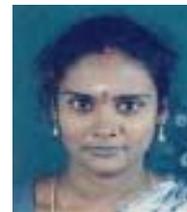